\def\year{2021}\relax
\documentclass[letterpaper]{article} % DO NOT CHANGE THIS
\usepackage{aaai21}  % DO NOT CHANGE THIS
\usepackage{times}  % DO NOT CHANGE THIS
\usepackage{helvet} % DO NOT CHANGE THIS
\usepackage{courier}  % DO NOT CHANGE THIS
\usepackage[hyphens]{url}  % DO NOT CHANGE THIS
\usepackage{graphicx} % DO NOT CHANGE THIS
\usepackage{natbib}  % DO NOT CHANGE THIS AND DO NOT ADD ANY OPTIONS TO IT
\usepackage{caption} % DO NOT CHANGE THIS AND DO NOT ADD ANY OPTIONS TO IT
\usepackage{mathtools}

\usepackage{multirow}

\usepackage[dvipsnames]{xcolor}
\usepackage{kotex}
\usepackage{booktabs}
\usepackage{amssymb}

\usepackage{subfig}
\usepackage{tabularx}

\usepackage{varwidth}

\usepackage{textcomp} % for microtype warning

\usepackage{floatrow}
\usepackage{booktabs}
\newfloatcommand{capbtabbox}{table}[][\FBwidth]
\usepackage[pagebackref=true,breaklinks=true,colorlinks,bookmarks=false,citecolor=black]{hyperref}

\newcommand{\ie}{\textit{i}.\textit{e}.}

\newcommand{\etal}{\textit{et} \textit{al}.}

\newcommand{\Fref}[1]{Fig.~\ref{#1}}

\newcommand{\Sref}[1]{Sec.~\ref{#1}}

\newcommand{\Tref}[1]{Table~\ref{#1}}

\newcommand{\R}{\mathbb{R}}

\newcommand{\abs}[1]{\left\lvert#1\right\rvert}
\newcommand{\norm}[1]{\left\lVert#1\right\rVert}

\title{Weakly-supervised Temporal Action Localization by Uncertainty Modeling}

\author{
    Pilhyeon Lee\textsuperscript{\rm 1}\thanks{Work done during his internship at Microsoft Research Asia.} \qquad Jinglu Wang\textsuperscript{\rm 2} \qquad
    Yan Lu\textsuperscript{\rm 2} \qquad
    Hyeran Byun\textsuperscript{\rm 1,3}\thanks{Corresponding author} \\
}
\affiliations {
    \textsuperscript{\rm 1} Department of Computer Science, Yonsei University \\
    \textsuperscript{\rm 2} Microsoft Research Asia\quad
    \textsuperscript{\rm 3} Graduate School of AI, Yonsei University\\
    \{lph1114, hrbyun\}@yonsei.ac.kr \quad
    \{jinglwa, yanlu\}@microsoft.com
}

\begin{document}

\maketitle

\begin{abstract}
Weakly-supervised temporal action localization aims to learn detecting temporal intervals of action classes with only video-level labels.
To this end, it is crucial to separate frames of action classes from the background frames (\ie, frames not belonging to any action classes).
In this paper, we present a new perspective on background frames where they are modeled as out-of-distribution samples regarding their inconsistency.
Then, background frames can be detected by estimating the probability of each frame being out-of-distribution, known as \textit{uncertainty}, but it is infeasible to directly learn uncertainty without frame-level labels. 
To realize the uncertainty learning in the weakly-supervised setting, we leverage the multiple instance learning formulation.
Moreover, we further introduce a background entropy loss to better discriminate background frames by encouraging their in-distribution (action) probabilities to be uniformly distributed over all action classes.
Experimental results show that our uncertainty modeling is effective at alleviating the interference of background frames and brings a large performance gain without bells and whistles.
We demonstrate that our model significantly outperforms state-of-the-art methods on the benchmarks, THUMOS'14 and ActivityNet (1.2 \& 1.3).
Our code is available at \url{https://github.com/Pilhyeon/WTAL-Uncertainty-Modeling}.
\end{abstract}

\section{Introduction}
\label{sec:intro}
Temporal action localization (TAL) is a very challenging problem of finding and classifying action intervals in untrimmed videos, which plays an important role in video understanding and analysis.
To tackle the problem, many works have been done in the fully-supervised manner and achieved impressive progress~\cite{Zeng2019GraphCN,Lin2019BMNBN,lin2020fast,Xu2020GTAD}.
However, they suffer from the extremely high cost of acquiring precise annotations, \ie, labeling the start and end timestamps of each action instance.

To relieve the high-cost issue and enlarge the scalability, researchers direct their attention to the same task with weak supervision, namely, weakly-supervised temporal action localization (WTAL).
Among the various levels of weak supervision, thanks to the cheap cost, video-level action labels are widely employed~\cite{wang2017untrimmednets}.
In this setting, each video is labeled as positive for action classes if it contains corresponding action frames and as negative otherwise.
Note that a video may have multiple action classes as its label.

\begin{figure}[t]
  \centering
  \includegraphics[clip=true, width=0.91\textwidth]{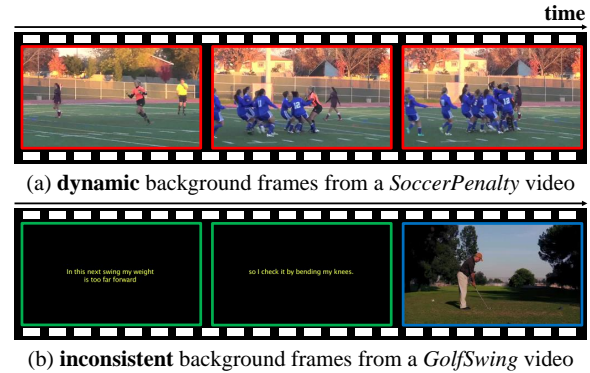}
  \caption{Challenges in modeling background: dynamism and inconsistency.
    (a)~The background frames in \textcolor{red}{red boxes} showing soccer players celebrating are very dynamic.
    (b)~The background frames are inconsistent. The frames in \textcolor{ForestGreen}{green boxes} show empty scenes with subtitles and the one in a \textcolor{blue}{blue box} shows a golfer preparing to shoot. 
    These two types do not have consistent appearances and semantics.}
  \label{fig:intro_fig}
\end{figure}

Existing approaches commonly cast WTAL as frame-wise classification, and adopt attention mechanism~\cite{nguyen2018weakly} or multiple instance learning~\cite{paul2018w} to learn from video-level labels.
Nonetheless, they still show highly inferior performances when compared to fully-supervised counterparts.
According to the literature~\cite{Xu2019SegregatedTA}, the performance degradation mainly comes from the false alarms of background frames, since video-level labels do not have any clue for background.
To bridge the gap, there appear several studies attempting explicit background modeling in the weakly-supervised manner.
Liu \etal~\cite{liu2019completeness} merges static frames to synthesize pseudo background videos, but they overlook dynamic background frames (\Fref{fig:intro_fig}a).
Meanwhile, some work~\cite{Nguyen2019WeaklySupervisedAL,lee2020background} tries to classify background frames as a separate class.
However, it is undesirable to force all background frames to belong to one specific class, as they do not share any common semantics (\Fref{fig:intro_fig}b).

\begin{figure}[t]%
    \centering
    \subfloat[Original features
    ]{{\includegraphics[clip=true, width=0.48\textwidth]{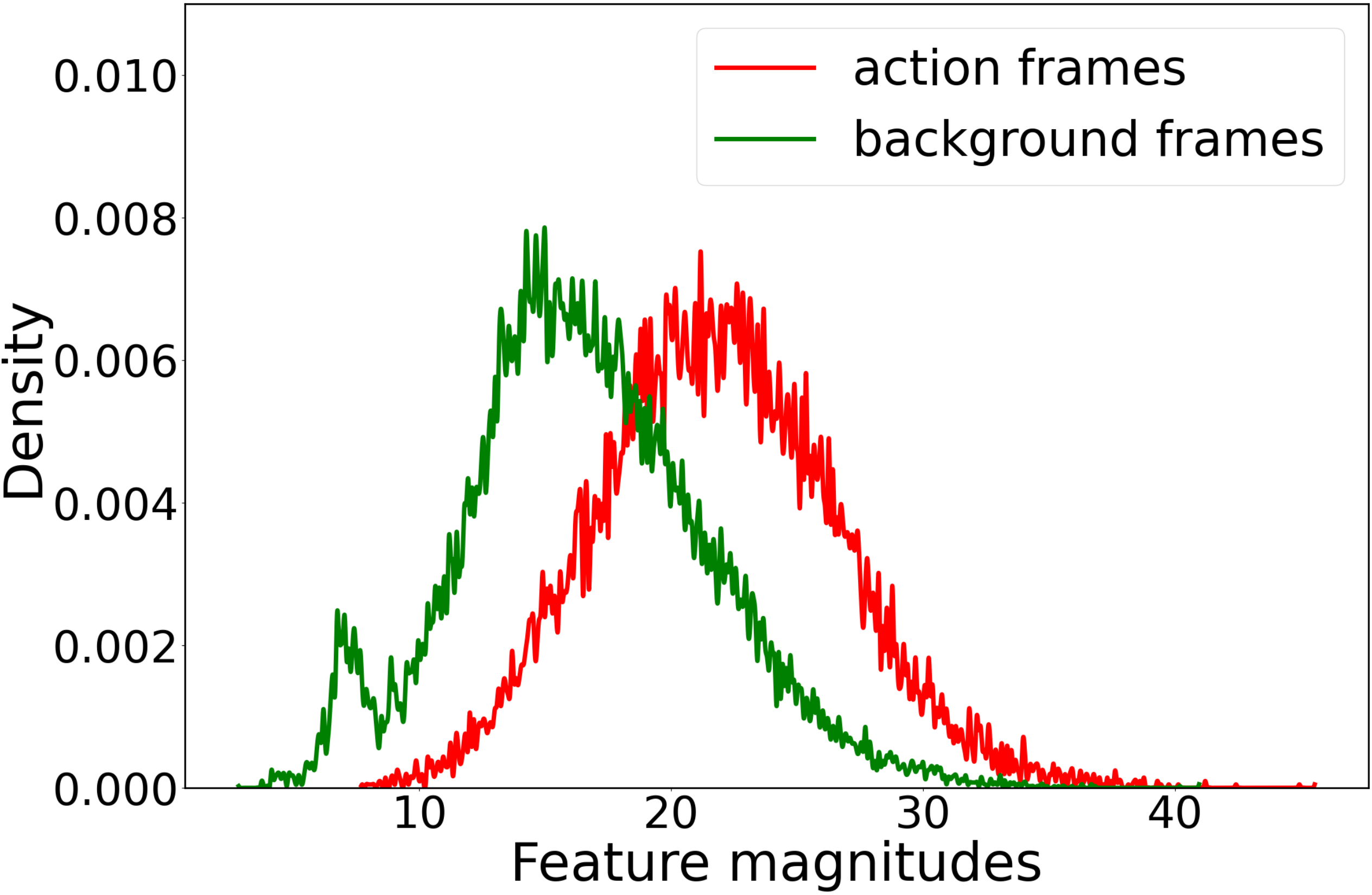}}
    \label{fig:magnitude_fig_1}}%
    \subfloat[Separated features
        ]{{\includegraphics[clip=true, width=0.47\textwidth]{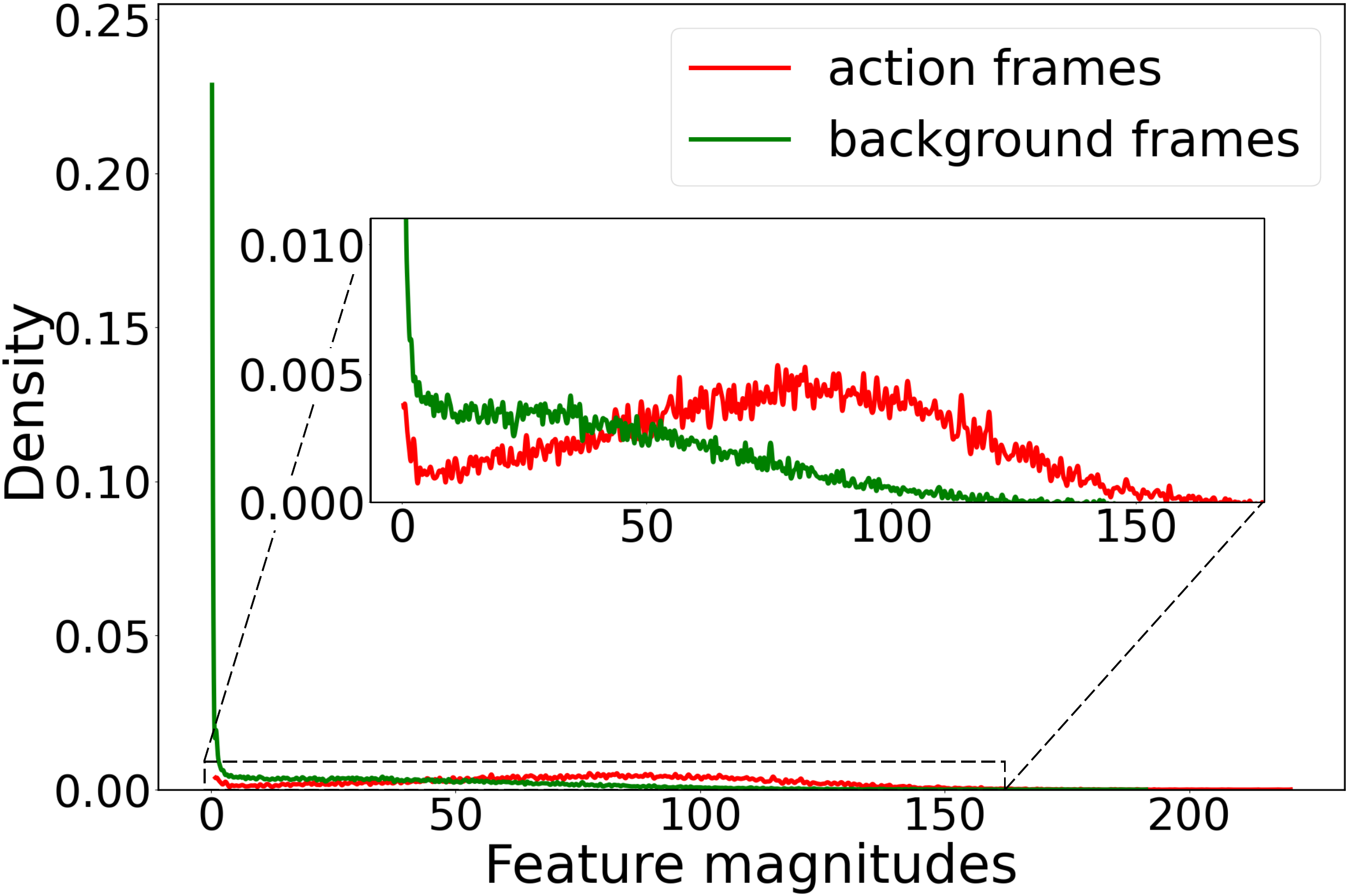}}
    \label{fig:magnitude_fig_2}}%
    \caption{
    Normalized histograms of feature magnitudes.
    (a)~The magnitudes of original features show correlation with action/background, but the centers of two distributions are close.
    (b)~The separated feature magnitudes by our method have better discrimination of action and background.
    }%
    \label{fig:magnitude_fig}%
\end{figure}

In this paper, we embrace the observation on inconsistency of background frames, and propose to formulate them as out-of-distribution samples~\cite{liang2018enhancing,Dhamija2018ReducingNA}.
They can be identified by estimating the probability of each sample coming from out-of-distribution, also known as \textit{uncertainty}~\cite{bendale2016towards,lakshminarayanan2017simple}.
To model uncertainty, we propose to utilize the magnitude of an embedded feature vector.
Generally, features of action frames have larger magnitudes compared to ones from background frames, as shown in \Fref{fig:magnitude_fig_1}.
This is because action frames need to produce high logits for ground-truth action classes.
Although the features magnitudes show the correlation with the discrimination between background and action frames, directly using them for discrimination is insufficient since the distributions of action and background are close to each other.
Therefore, to further encourage the discrepancy in feature magnitudes, we propose to separate the distributions by enlarging magnitudes of action features and reducing those of background features close to zero (\Fref{fig:magnitude_fig_2}).

In order to learn uncertainty only with video-level supervision, we leverage the formulation of multiple instance learning~\cite{Maron1998MultipleInstanceAF,zhou2004multi}, where a model is trained with a bag (\ie, untrimmed video) instead of instances (\ie, frames).
Specifically, from each untrimmed video, we select top-$k$ and bottom-$k$ frames based on the feature magnitude and consider them as pseudo action and background frames, respectively.
Thereafter, we design an uncertainty modeling loss to separate their magnitudes,
through which our model is able to indirectly model uncertainty without frame-level labels and provides better separation between action and background frames.
Moreover, we introduce a background entropy loss to force pseudo background frames to have uniform probability distribution over action classes.
This prevents them from leaning toward a certain action class and helps to reject them by maximizing the entropy of their action class distribution.
To validate the effectiveness of our method, we perform experiments on two standard benchmarks, THUMOS'14 and ActivityNet.
By jointly optimizing the proposed losses along with a general action classification loss, our model successfully distinguishes action frames from background frames.
Furthermore, our method achieves the new state-of-the-art performances on the both benchmarks.

Our contributions are three-fold:
1) We propose to formulate background frames as out-of-distribution samples, overcoming the difficulty in modeling background due to their inconsistency. 
2) We design a new framework for weakly-supervised action localization, where uncertainty is modeled and learned only with video-level labels via multiple instance learning.
3) We further encourage separation between action and background with a loss maximizing the entropy of action probability distribution from background frames.

\section{Related Work}
\paragraph{Fully-supervised action localization.}
The goal of temporal action localization is to find temporal intervals of action instances from long untrimmed videos and classify them. For the task, many approaches depend on accurate temporal annotations for each training video.
Most of them utilize the two-stage approach, \ie, they first generate proposals, and then classify them.
To generate proposals, earlier methods adopt the sliding window technique~\cite{shou2016temporal,yuan2016temporal,shou2017cdc,yang2018exploring,xiong2017pursuit,chao2018rethinking}, while recent models predict the start and end frames of action~\cite{lin2018bsn,Lin2019BMNBN,lin2020fast}.
Meanwhile, there appear some attempts to leverage graph structural information~\cite{Zeng2019GraphCN,Xu2020GTAD}.
Moreover, there are also a gaussian modeling of each action instance~\cite{long2019gaussian} and an efficient method without proposal generation step~\cite{alwassel2018action}.

\paragraph{Weakly-supervised action localization.}
Recently, many attempts have been made to solve temporal action localization with weak supervision, mostly video-level labels.
\cite{wang2017untrimmednets} first tackle the problem by selecting relevant segments on soft and hard ways. 
    \cite{nguyen2018weakly} propose a sparsity regularization, while \cite{singh2017hide} and \cite{Yuan2019MARGINALIZEDAA} extend small discriminative parts.
\cite{paul2018w}, \cite{Narayan20193CNetCC} and \cite{min2020A2CL} employ deep metric learning to force features from the same action to get closer to themselves than those from different classes.
\cite{Shi2020DGAM} and \cite{luo2020EMMIL} learn attention weights using a variational auto-encoder and an expectation-maximization strategy, respectively.
\cite{zhai2020TSCN} pursue the consensus between different modalities.
Meanwhile, \cite{shou2018autoloc} and \cite{Liu2019WeaklyST} attempt to regress the intervals of action instances, instead of performing hard thresholding.
Recently, \cite{ma2020sfnet} propose to exploit the intermediate level of supervision (\ie, single frame supervision).

\begin{figure*}[t]
  \centering
  \includegraphics[clip=true, width=0.93\textwidth]{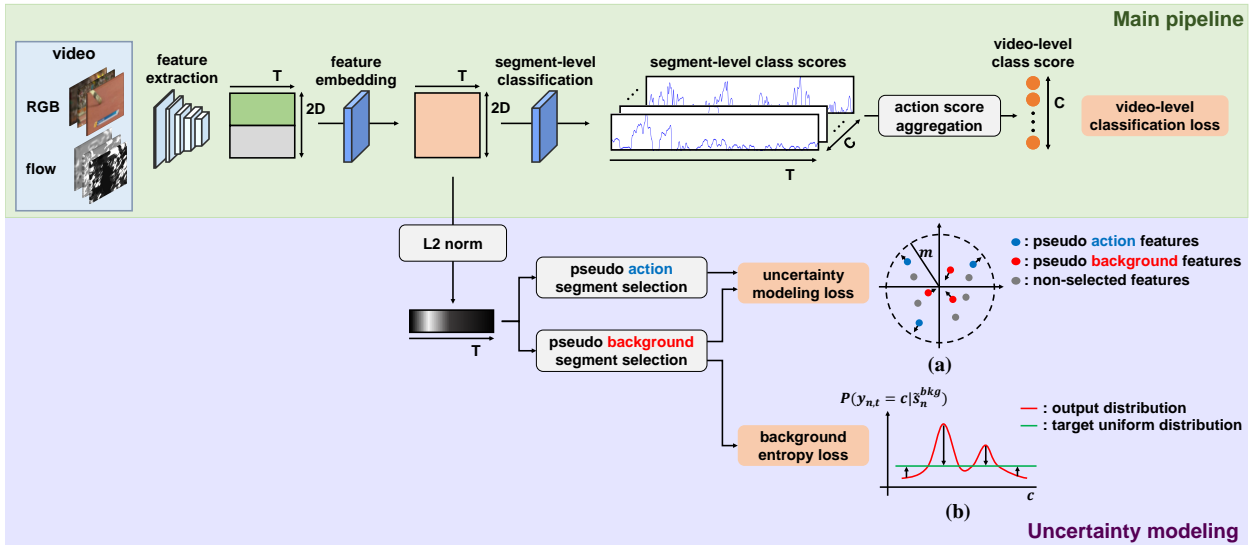}
  \caption{Overview of the proposed method.
  The main pipeline serves the standard process for weakly-supervised action localization. We discriminate background frames from action frames by modeling \textit{uncertainty}, \ie, probability of being out-of-distribution.
  In the uncertainty modeling part, 
  pseudo action/background segments are selected based on features magnitudes, which are in turn used to derive two proposed losses for background discrimination: (a) uncertainty modeling loss which enlarges and reduces the feature magnitudes of the pseudo action and background segments respectively; and (b) background entropy loss forcing the pseudo background segments to have uniform action probability distribution.}
  \label{fig:main_architecture}
\end{figure*}

Apart from the methods above, Some work \cite{liu2019completeness,Nguyen2019WeaklySupervisedAL,lee2020background} seeks to explicitly model background. However, as aforementioned, they have innate limitations in that background frames could be dynamic and inconsistent.
In contrast, we consider background as out-of-distribution and propose to learn uncertainty as well as action class scores.
In \Sref{sec:experiments}, the efficacy of our approach is verified.

\section{Method}
\label{sec:method}
In this section, we first set up the baseline network for weakly-supervised temporal action localization~(\Sref{subsec:main_pipe}).
Thereafter, we cast the problem of identifying background frames as out-of-distribution detection and tackle it by modeling uncertainty (\Sref{subsec:ood_formul}).
Lastly, the objective functions for training our model~(\Sref{subsec:optimization}) and how the inference is performed~(\Sref{subsec:inference}) are elaborated.
The overview of our method is illustrated in \Fref{fig:main_architecture}.

\subsection{Main pipeline}
\label{subsec:main_pipe}
\paragraph{Feature extraction.}
Due to the memory limit, we split each video into multi-frame segments, \ie, $v_{n}=\{s_{n,l}\}_{l=1}^{L_{n}}$, where $L_{n}$ denotes the number of segments in $v_{n}$.
To handle the large variation in video lengths, a fixed number of $T$ segments $\{\tilde{s}_{n,t}\}_{t=1}^{T}$ are sampled from each original video.
From the sampled RGB and flow segments, spatio-temporal features $x_{n,t}^{\text{RGB}}\in\R^{D}$ and $x_{n,t}^{\text{flow}}\in\R^{D}$ are extracted by a pre-trained feature extractor, respectively.
Next, we concatenate the RGB and flow features into the complete feature vectors $x_{n,t}\in\R^{2D}$, which are then stacked to build a feature map of length $T$, \ie, $X_{n}=\begin{bmatrix}x_{n,1},...,x_{n,T}\end{bmatrix}\in\R^{2D \times T}$.

\paragraph{Feature embedding.} 
In order to embed the extracted features into the task-specific space, we employ a single 1-D convolutional layer followed by a ReLU function. Formally, $F_{n}= g_{\text{embed}}(X_{n};\phi_{\text{embed}})$, where $g_{\text{embed}}$ denotes the convolution operator with the activation and $\phi_{\text{embed}}$ is the trainable parameters of the layer.
Concretely, the dimension of an embedded feature is the same as that of an input feature, \ie, $F_{n}=\begin{bmatrix}f_{n,1},...,f_{n,T}\end{bmatrix}\in\R^{2D \times T}$.

\paragraph{Segment-level classification.}
From the embedded features, we predict segment-level class scores, which are later used for action localization.
Given the features $F_{n}$, the segment-wise class scores are derived by the action classifier, \ie, $\mathcal{A}_{n} = g_{\text{cls}}(F_{n};\phi_{\text{cls}})$,
where $g_{\text{cls}}$ represents the linear classifier with its parameters $\phi_{\text{cls}}$.
In specific, $\mathcal{A}_{n}\in\R^{C \times T}$, where $C$ is the number of action classes.

\paragraph{Action score aggregation.}
Following the previous work \cite{paul2018w},
we aggregate top $k^{\text{act}}$ scores along all segments for each action class and average them to build a video-level raw class score:
\begin{equation}
    \label{equ:aggregation}
    a_{c}(v_{n})=\frac{1}{k^{\text{act}}}\max_{\substack{\hat{\mathcal{A}}_{n;c}\subset\mathcal{A}_{n}[c,:]}} \sum_{\forall a \in \hat{\mathcal{A}}_{n;c}}{a},
\end{equation}
where $\hat{\mathcal{A}}_{n;c}$ is a subset containing $k^{\text{act}}$ action scores for the $c$-th class (\ie, $\abs{\hat{\mathcal{A}}_{n;c}}=k^{\text{act}}$), and
$k^{\text{act}}$ is a hyper-parameter controlling the number of the aggregated segments.
Thereafter, we obtain the video-level action probability for each action class by applying the softmax function to the aggregated scores:
\begin{equation}
    \label{equ:video_level_score}
    p_{c}(v_{n})=\frac{\text{exp}(a_{c}(v_{n}))}{\sum_{c'=1}^{C}\text{exp}(a_{c'}(v_{n}))},
\end{equation}
where $p_{c}(v_{n})$ represents the softmax score for the $c$-th action of $v_{n}$, which is guided by video-level weak labels.

\subsection{Uncertainty modeling}
\label{subsec:ood_formul}
From the main pipeline, we can obtain the action probabilities for each segment, but the essential component for action localization, \ie, background discrimination, is not carefully considered.
Regarding the unconstraint and inconsistency of background frames, we treat background as out-of-distribution~\cite{hendrycks2016baseline} and model \textit{uncertainty} (probability of being background) for WTAL.

Considering the probability that the segment $\tilde{s}_{n,t}$ belongs to the $c$-th action, it can be decomposed into two parts with a chain rule, \ie, the in-distribution action probability and the uncertainty. 
Let $d \in \{0,1\}$ denote the variable for the background discrimination, \ie, $d=1$ if the segment belongs to any action class, $d=0$ otherwise (belongs to background).
Then, the posterior probability for class $c$ of $\tilde{s}_{n,t}$ is given by:
\begin{equation}
\small
\label{equ:posterior}
\begin{aligned}
    P(y_{n,t}=c|\tilde{s}_{n,t})& = P(y_{n,t}=c, d=1|\tilde{s}_{n,t}) \\
            & = P(y_{n,t}=c|d=1, \tilde{s}_{n,t}) P(d=1|\tilde{s}_{n,t}),
\end{aligned}
\end{equation}
where $y_{n,t}$ is the label of the segment, \ie, if $\tilde{s}_{n,t}$ belongs to the $c$-th action class, then $y_{n,t}=c$, while $y_{n,t}=0$ for background segments.
We describe the single-label case for readability. Without loss of generality, this can be generalized to the multi-label.

\paragraph{Uncertainty formulation.}
In Eq.~\ref{equ:posterior}, the probability for in-distribution action classification, $P(y_{n,t}=c|d=1, \tilde{s}_{n,t})$, is estimated with the softmax function as in general classification task.
Additionally, it is necessary to model the probability that a segment belongs to any action class, \ie, $P(d=1|\tilde{s}_{n,t})$, to tackle the background discrimination problem.
Observing that features of action frames generally have larger  magnitudes than those of background frames (\Fref{fig:magnitude_fig}), we formulate uncertainty by employing magnitudes of feature vectors.
Specifically,
background features have small magnitudes close to 0, while action features have large ones.
Then the probability that the $t$-th segment in the $n$-th video ($\tilde{s}_{n,t}$) is an action segment is defined by:
\begin{equation}
\label{equ:ood_formul}
    P(d=1\mathbf{|}\tilde{s}_{n,t}) = \frac{\text{min}(m, \norm{f_{n,t}})}{m},
\end{equation}
where $f_{n,t}$ is the corresponding feature vector of $\tilde{s}_{n,t}$, $\norm{\cdot}$ is a norm function (we use L-2 norm here), and $m$ is the pre-defined maximum feature magnitude. From the formulation, it is ensured that the probability falls between 0 and 1, \ie, $0 \leq P(d=1\mathbf{|}\tilde{s}_{n,t}) \leq 1$.

\paragraph{Uncertainty learning via multiple instance learning.}
To learn uncertainty only with video-level labels, we borrow the concept of multiple instance learning~\cite{Maron1998MultipleInstanceAF}, where a model is trained with a bag (video) instead of instances (segments).
In this setting, considering that each untrimmed video contains both action and background frames, we select pseudo action/background segments representing the video.
Specifically, the top $k^{\text{act}}$ segments in terms of the feature magnitude are treated as the pseudo action segments $\{\tilde{s}_{n,i}|i\in\mathcal{S}^{\text{act}}\}$, where $\mathcal{S}^{\text{act}}$ indicates the set of pseudo action indices.
Meanwhile, the bottom $k^{\text{bkg}}$ segments are considered the pseudo background segments $\{\tilde{s}_{n,j}|j\in\mathcal{S}^{\text{bkg}}\}$, where $\mathcal{S}^{\text{bkg}}$ denotes the set of indices for pseudo background.
% $|\mathcal{S}^{\text{act}}|=k^{\text{act}}$ and $|\mathcal{S}^{\text{bkg}}|=k^{\text{bkg}}$, where
$k^{\text{act}}$ and $k^{\text{bkg}}$ represent the number of segments sampled for action and background, respectively. 
Then the pseudo action/background segments serve as the representatives of the input untrimmed video, and they are used for training via multiple instance learning.

\subsection{Training objectives}
\label{subsec:optimization}
Our model is jointly optimized with three losses: 1) video-level classification loss $\mathcal{L}_{\text{cls}}$ for action classification of each input video, 2) uncertainty modeling loss $ \mathcal{L}_{\text{um}}$ for separating the magnitudes of action and background feature vectors, and 3) background entropy loss $\mathcal{L}_{\text{be}}$ which forces background segments to have uniform probability distribution for action classes. The overall loss function is as follows:
\begin{equation}
  \label{equ:loss_total}
  \mathcal{L}_{\text{total}} = \mathcal{L}_{\text{cls}}+\alpha \mathcal{L}_{\text{um}}+ \beta \mathcal{L}_{\text{be}},
\end{equation}
where $\alpha$ and $\beta$ are balancing hyper-parameters.

\paragraph{Video-level classification loss.}
For multi-label action classification, we use the binary cross entropy loss with normalized video-level labels~\cite{wang2017untrimmednets} as follows:
\begin{equation}
  \label{equ:loss_video}
  \mathcal{L}_{\text{cls}} = \frac{1}{N}\sum_{n=1}^{N}\sum_{c=1}^{C}-y_{n;c}\log p_{c}(v_{n}),
\end{equation}
where $p_{c}(v_{n})$ represents the video-level softmax score for the $c$-th class of the $n$-th video (Eq. \ref{equ:video_level_score}), and $y_{n;c}$ is the normalized video-level label for the $c$-th class of the $n$-th video.

\paragraph{Uncertainty modeling loss.}
To learn uncertainty, we train the model to produce large feature magnitudes for pseudo action segments but ones with small ones for pseudo background segments, as illustrated in \Fref{fig:main_architecture} (a).
Formally, uncertainty modeling loss takes the form:

\begin{equation}
  \label{equ:loss_dist}
  \mathcal{L}_{\text{um}} = \frac{1}{N}\sum_{n=1}^{N}(\max(0,m-\norm{f_{n}^{\text{act}}})+\norm{f_{n}^{\text{bkg}}})^{2},
\end{equation}
where $f_{n}^{\text{act}}=\frac{1}{k^{\text{act}}}\sum_{i \in \mathcal{S}^{\text{act}}}f_{n,i}$ and $f_{n}^{\text{bkg}}=\frac{1}{k^{\text{bkg}}}\sum_{j \in \mathcal{S}^{\text{bkg}}}f_{n,j}$ are the mean features of the pseudo action and background segments of the $n$-th video, respectively.
$\norm{\cdot}$ is the norm function, and $m$ is the pre-defined maximum feature magnitude, the same in Eq. \ref{equ:ood_formul}.

\paragraph{Background entropy loss.}
Although uncertainty modeling loss encourages background segments to produce low logits for all actions, softmax scores for some action classes could be high due to the relativeness of softmax function.
To prevent background segments from having a high softmax score for any action class, we define a loss function which maximizes the entropy of action probability from background segments, \ie, background segments are forced to have uniform probability distribution for action classes as described in \Fref{fig:main_architecture} (b). The loss is calculated as follows:

\begin{equation}
  \label{equ:loss_entropy}
  \mathcal{L}_{\text{be}} = \frac{1}{NC}\sum_{n=1}^{N}\sum_{c=1}^{C}-\log(p_{c}(\tilde{s}_{n}^{\text{bkg}})),
\end{equation}
where $p_{c}(\tilde{s}_{n}^{\text{bkg}})=\frac{1}{k^{\text{bkg}}}\sum_{j \in \mathcal{S}^{\text{bkg}}}p_{c}(\tilde{s}_{n,j})$ is the averaged action probability for the $c$-th class of pseudo background segments, and
$p_{c}(\tilde{s}_{n,j})$ is the softmax score for class $c$ of $\tilde{s}_{n,j}$.

\begin{table*}[t]
\caption{
Comparison on THUMOS'14. AVG is the average mAP under the thresholds 0.1:0.1:0.7, while ${\dagger}$ indicates the use of additional information, such as action frequency or human pose.
}
\begin{center}
\resizebox{.9\textwidth}{!}{
\begin{tabular}{c|l|ccccccc|c}
\toprule
\multirow{2}{*}{Supervision}       & \multicolumn{1}{c|}{\multirow{2}{*}{Method}} &
\multicolumn{8}{c}{mAP@IoU (\%)}         \\
       & \multicolumn{1}{c|}{}  & 0.1   & 0.2   & 0.3   & 0.4   & 0.5   & 0.6   & 0.7   & AVG  \\
      \midrule\midrule
\multirow{6}{*}{Full}
      & S-CNN~\cite{shou2016temporal}      & 47.7  & 43.5  & 36.3  & 28.7  & 19.0  & 10.3  & 5.3  & 27.3  \\
      & SSN~\cite{zhao2017temporal}      & 66.0  & 59.4  & 51.9  & 41.0  & 29.8  & -     & -     & -    \\
       & TAL-Net~\cite{chao2018rethinking}     & 59.8  & 57.1   & 53.2  & 48.5                     & 42.8                     & 33.8                     & 20.8                     & 45.1                    \\
       & BSN~\cite{lin2018bsn}       & -  & -  & 53.5  & 45.0  & 36.9  & 28.4  & 20.0  & - \\
       & P-GCN~\cite{Zeng2019GraphCN}       & 69.5                     & 67.8  & 63.6  & 57.8  & 49.1  & -  & -  & - \\
       & G-TAD~\cite{Xu2020GTAD}       & -  & -  & 66.4  & 60.4  & 51.6  & 37.6  & 22.9  & - \\
        \midrule
\multirow{3}{*}{Weak${\dagger}$}
        & STAR~\cite{Xu2019SegregatedTA}  & 68.8  & 60.0  & 48.7  & 34.7  & 23.0   & -  & -  & -    \\
        & 3C-Net~\cite{Narayan20193CNetCC}  & 59.1  & 53.5  & 44.2  & 34.1  & 26.6   & -  & 8.1  & -    \\
        & PreTrimNet~\cite{zhang2020MultiinstanceMA}  & 57.5  & 50.7  & 41.4  & 32.1  & 23.1  & 14.2  & 7.7  & 23.7    \\
        \midrule
\multirow{19}{*}{Weak} 
      & UntrimmedNets~\cite{wang2017untrimmednets}      & 44.4  & 37.7  & 28.2  & 21.1  & 13.7  & -     & -  & -    \\
      & Hide-and-seek~\cite{singh2017hide}     & 36.4  & 27.8  & 19.5  & 12.7  & 6.8   & -     & -  & -    \\
      & STPN~\cite{nguyen2018weakly}    & 52.0  & 44.7  & 35.5  & 25.8  & 16.9  & 9.9   & 4.3   & 27.0     \\
      & AutoLoc~\cite{shou2018autoloc}      & -  & -  & 35.8  & 29.0  & 21.2  & 13.4  & 5.8  & -  \\
      & W-TALC~\cite{paul2018w}      & 55.2  & 49.6  & 40.1  & 31.1  & 22.8  & -     & 7.6  & -  \\
      & MAAN~\cite{Yuan2019MARGINALIZEDAA}      & 59.8  & 50.8   & 41.1  & 30.6  & 20.3  & 12.0  & 6.9  & 31.6  \\
      & Liu \textit{et al.}~\cite{liu2019completeness}      & 57.4  & 50.8  & 41.2  & 32.1  & 23.1  & 15.0  & 7.0  & 32.4  \\
      & CleanNet~\cite{Liu2019WeaklyST}            & - & - & 37.0 & 30.9 & 23.9 & 13.9 & 7.1  & -  \\
      & TSM~\cite{Yu2019TemporalSM}      & -  & -  & 39.5  & -  & 24.5  & -  & 7.1  & -  \\
      & Nguyen \textit{et al.}~\cite{Nguyen2019WeaklySupervisedAL}            & 60.4 & 56.0 & 46.6 & 37.5 & 26.8 & 17.6 & 9.0  & 36.3  \\
      & BaS-Net~\cite{lee2020background}            & 58.2 & 52.3 & 44.6 & 36.0 & 27.0 & 18.6 & 10.4  & 35.3  \\
      & RPN~\cite{huang2020relational}            & 62.3 & 57.0 & 48.2 & 37.2 & 27.9 & 16.7 & 8.1  & 36.8  \\
      & DGAM~\cite{Shi2020DGAM}            & 60.0 & 54.2 & 46.8 & 38.2 & 28.8 & 19.8 & 11.4  & 37.0  \\
      & Gong \textit{et al.}~\cite{Gong2020coattention}            & - & - & 46.9 & 38.9 & 30.1 & 19.8 & 10.4  & -  \\
      & ActionBytes~\cite{Jain2020ActionBytes}            & - & - & 43.0 & 35.8 & 29.0 & - & 9.5  & -  \\
      & EM-MIL~\cite{luo2020EMMIL}            & 59.1 & 52.7 & 45.5 & 36.8 & 30.5 & 22.7 & \textbf{16.4}  & 37.7  \\
      & A2CL-PT~\cite{min2020A2CL}            & 61.2 & 56.1 & 48.1 & 39.0 & 30.1 & 19.2 & 10.6  & 37.8  \\
      & TSCN~\cite{zhai2020TSCN}            & 63.4 & 57.6 & 47.8 & 37.7 & 28.7 & 19.4 & 10.2  & 37.8  \\
      \cmidrule(lr){2-10}
      & Ours            & \textbf{67.5} & \textbf{61.2} & \textbf{52.3} & \textbf{43.4} & \textbf{33.7} & \textbf{22.9} & 12.1  & \textbf{41.9}  \\
       \bottomrule
\end{tabular}
}
\end{center}
\label{table:quant_thumos}
\end{table*}

\subsection{Inference}
\label{subsec:inference}

At the test time, for an input video, we first obtain the video-level softmax score and threshold on it with $\theta_{\text{vid}}$ to determine which action classes are to be localized.
For the remaining action classes, we calculate the segment-wise posterior probability by multiplying the segment-level softmax score and the probability of being an action segment following Eq.~\ref{equ:posterior}. Afterwards, the segments whose posterior probabilities are larger than $\theta_{\text{seg}}$ are selected as the candidate segments.
Finally, consecutive candidate segments are grouped into a single proposal, whose scores are calculated following \cite{liu2019completeness}.
To enrich the proposal pool, we use multiple thresholds for $\theta_{\text{seg}}$ and non-maxium suppression (NMS) is performed for overlapped proposals.

\section{Experiments}
\label{sec:experiments}

\subsection{Experimental settings}

\paragraph{Datasets.}
THUMOS'14~\cite{THUMOS14} is a widely used dataset for temporal action localization, containing 200 validation videos and 213 test videos of 20 action classes. It is a very challenging benchmark as the lengths of the videos are diverse and actions frequently occur (on average 15 instances per video). Following the previous work, we use validation videos for training and test videos for test.
On the other hand, ActivityNet~\cite{caba2015activitynet} is a large-scale benchmark with two versions. ActivityNet 1.3, consisting of 200 action categories, includes 10,024 training videos, 4,926 validation videos and 5,044 test videos. ActivityNet 1.2 is a subset of the version 1.3, and is composed of 4,819 training videos, 2,383 validation videos and 2,480 test videos of 100 action classes. Because the ground-truths for the test videos of ActivityNet are withheld for the challenge, we utilize validation videos for evaluation.

\paragraph{Evaluation metrics.}
We evaluate our method with mean average precisions (mAPs) under several different intersection of union (IoU) thresholds, which are the standard evaluation metrics for temporal action localization. The official evaluation code of ActivityNet\footnote{\url{https://github.com/activitynet/ActivityNet}} is used for measurement.

\paragraph{Implementation details.}
As the feature extractor, we employ I3D networks~\cite{carreira2017quo} pre-trained on Kinetics~\cite{carreira2017quo}, which take input segments with 16 frames.
It should be noted that we do not finetune the feature extractor for fair comparison.
TVL1 algorithm~\cite{wedel2009improved} is used to extract optical flow from videos. We fix the number of segments $T$ as 750 and 50 for THUMOS'14 and ActivityNet, respectively.
The sampling method is the same as STPN~\cite{nguyen2018weakly}.
The number of the pseudo action/background frames is determined by the ratio parameters, \ie, $k^{\text{act}}=T/r^{\text{act}}$ and $k^{\text{bkg}}=T/r^{\text{bkg}}$.
All hyper-parameters are set by grid search; $m=100$, $r^{\text{act}}=9$, $r^{\text{bkg}}=4$, $\alpha=5\times10^{-4}$, $\beta=1$, and $\theta_{\text{vid}}=0.2$.
Multiple thresholds from 0 to 0.25 with a step size 0.025 are used as $\theta_{\text{seg}}$, then we perform non-maxium suppression (NMS) with an IoU threshold of 0.6.

\begin{table}[t]
\caption{
Results on ActivityNet 1.2. AVG is the averaged mAP at the thresholds 0.5:0.05:0.95, while ${\dagger}$ means the use of action counts.
}
\begin{center}
\resizebox{0.87\columnwidth}{!}{
\begin{tabular}{c|l|ccc|c}
\toprule
\multirow{2}{*}{Sup.} &
\multicolumn{1}{c|}{\multirow{2}{*}{Method}} &
\multicolumn{4}{c}{mAP@IoU (\%)}  \\ &
\multicolumn{1}{c|}{} & 0.5  & 0.75  & 0.95  & AVG  \\
      \midrule\midrule
\multirow{1}{*}{Full}
      & SSN~\shortcite{zhao2017temporal}  & 41.3  & 27.0  & 6.1  & 26.6    \\
      \midrule
\multirow{1}{*}{Weak${\dagger}$}
        & 3C-Net~\shortcite{Narayan20193CNetCC}  & 37.2  & -  & -  & 21.7  \\
        \midrule
\multirow{13}{*}{Weak} 
       & UntrimmedNets~\shortcite{wang2017untrimmednets}  & 7.4  & 3.2  & 0.7  & 3.6  \\
       & AutoLoc~\shortcite{shou2018autoloc}  & 27.3  & 15.1  & 3.3  & 16.0  \\
       & W-TALC~\shortcite{paul2018w}  & 37.0  & 12.7  & 1.5  & 18.0  \\
       & Liu \textit{et al.}~\shortcite{liu2019completeness}  & 36.8  & 22.9  & 5.6  & 22.4  \\
       & CleanNet~\shortcite{Liu2019WeaklyST}  & 37.1  & 20.3  & 5.0  & 21.6  \\
       & TSM~\shortcite{Yu2019TemporalSM}  & 28.3  & 17.0  & 3.5  & 17.1  \\
       & RPN~\shortcite{huang2020relational}   & 37.6  & 23.9  & 5.4  & 23.3  \\
       & BaS-Net~\shortcite{lee2020background}  & 38.5  & 24.2  & 5.6  & 24.3  \\
       & DGAM~\shortcite{lee2020background}  & 41.0  & 23.5  & 5.3  & 24.4  \\
       & Gong \textit{et al.}~\shortcite{Gong2020coattention}  & 40.0  & 25.0  & 4.6  & 24.6  \\
       & EM-MIL~\shortcite{luo2020EMMIL}  & 37.4  & -  & -  & 20.3  \\
       & TSCN~\shortcite{zhai2020TSCN}  & 37.6  & 23.7  & 5.7  & 23.6  \\
       \cmidrule(lr){2-6}
       & Ours  & \textbf{41.2}  & \textbf{25.6}  & \textbf{6.0}  & \textbf{25.9}  \\
\bottomrule

\end{tabular}
}
\end{center}
\label{table:quant_anet12}
\end{table}

\begin{table}[t]
\caption{
Ablation study on THUMOS'14. AVG represents the average mAP at IoU thresholds 0.1:0.1:0.7.
``Score'' indicates the way of calculating the final scores.
To derive final scores, the softmax method uses the segment-level softmax scores, while the fusion method fuses the segment-level softmax scores and uncertainty as in Eq.~\ref{equ:posterior}.
}
\begin{center}
\resizebox{1.0\textwidth}{!}{
\begin{tabular}{cc|cc|ccccccc|c}
\toprule
\multicolumn{2}{c|}{Score}
& \multicolumn{2}{c|}{Loss}
& \multicolumn{8}{c}{mAP@IoU (\%)}  \\
\cmidrule(lr){1-2}
\cmidrule(lr){3-4}
% \cline{1-4]}
softmax  & fusion  & $\mathcal{L}_{\text{um}}$  & $\mathcal{L}_{\text{be}}$  & 0.1  & 0.2  & 0.3  & 0.4  & 0.5  & 0.6  & 0.7  & AVG  \\
\midrule\midrule
\checkmark  &  &  &  & 42.3  & 35.1  & 27.7  & 21.7  & 16.4  & 10.5  & 4.6  & 22.6  \\
& \checkmark  &   &  & 58.8  & 50.0  & 40.5  & 31.7  & 23.6  & 15.5  & 7.2  & 32.5  \\
& \checkmark  & \checkmark  &  & 65.3  & 59.2  & 50.8  & 42.3  & 32.4  & 21.1  & 10.4  & 40.2  \\
& \checkmark  & \checkmark  & \checkmark  & 67.5  & 61.2  & 52.3  & 43.4  & 33.7  & 22.9  & 12.1  & 41.9  \\
\bottomrule
    \end{tabular}
}
\end{center}
\label{table:ablation_thumos_1}
\end{table}

\begin{table}[t]
\caption{
Comparison on ActivityNet 1.3. $\dagger$ means the use of external data and AVG is the averaged mAP under the thresholds 0.5:0.05:0.95.
}
\begin{center}
\resizebox{.85\columnwidth}{!}{
\begin{tabular}{c|l|ccc|c}
\toprule
\multirow{2}{*}{Sup.} &
\multicolumn{1}{c|}{\multirow{2}{*}{Method}} &
\multicolumn{4}{c}{mAP@IoU (\%)}  \\ &
\multicolumn{1}{c|}{}  & 0.5  & 0.75 & 0.95  & AVG  \\
      \midrule\midrule
\multirow{5}{*}{Full}
      & TAL-Net~\shortcite{chao2018rethinking}  & 38.2  & 18.3  & 1.3  & 20.2    \\
      & BSN~\shortcite{lin2018bsn}  & 46.5  & 30.0  & 8.0  & 30.0    \\
      & BMN~\shortcite{Lin2019BMNBN}  & 50.1  & 34.8  & 8.3  & 33.9  \\
      & P-GCN~\shortcite{Zeng2019GraphCN}  & 48.3  & 33.2  & 3.3  & 31.1    \\
      & G-TAD~\shortcite{Xu2020GTAD}  & 50.4  & 34.6  & 9.0  & 34.1    \\
      \midrule
\multirow{2}{*}{Weak${\dagger}$}
        & STAR~\shortcite{Xu2019SegregatedTA}  & 31.1  & 18.8  & 4.7  & -  \\
        & PreTrimNet~\shortcite{zhang2020MultiinstanceMA}  & 34.8  & 20.9  & 5.3  & 22.5  \\ \midrule
\multirow{9}{*}{Weak} 
       & STPN~\shortcite{nguyen2018weakly}  & 29.3  & 16.9  & 2.6  & -  \\
       & MAAN~\shortcite{Yuan2019MARGINALIZEDAA}  & 33.7  & 21.9  & 5.5  & -  \\
       & Liu \textit{et al.}~\shortcite{liu2019completeness}  & 34.0  & 20.9  & \textbf{5.7}  & 21.2  \\
       & TSM~\shortcite{Yu2019TemporalSM}  & 30.3  & 19.0  & 4.5  & -  \\
       & Nguyen \textit{et al.}~\shortcite{Nguyen2019WeaklySupervisedAL}  & 36.4  & 19.2  & 2.9  & -  \\
       & BaS-Net~\shortcite{lee2020background}  & 34.5  & 22.5  & 4.9  & 22.2  \\
       & A2CL-PT~\shortcite{min2020A2CL}  & 36.8  & 22.5  & 5.2  & 22.5  \\
       & TSCN~\shortcite{zhai2020TSCN}  & 35.3  & 21.4  & 5.3  & 21.7  \\
      \cmidrule(lr){2-6}
       & Ours  & \textbf{37.0}  & \textbf{23.9}  & \textbf{5.7}  & \textbf{23.7}  \\
\bottomrule
\end{tabular}
}
\end{center}
\label{table:quant_anet13}
\end{table}

\begin{table}[t]
\caption{
Analysis on the maximum feature magnitude $m$ on THUMOS'14. We report the average mAPs under IoU thresholds 0.1:0.1:0.7 with varying $m$ from 10 to 250.
}
\begin{center}
\resizebox{.87\textwidth}{!}{
\begin{tabular}{c|ccccccc}
\toprule
$m$ & 10 & 25 & 50 & 100 & 150 & 200 & 250\\
\midrule\midrule
mAP@AVG & 39.6 & 41.1 & 41.8 & 41.9  & 41.7  & 41.9  & 41.6 \\
\bottomrule
    \end{tabular}
}
\end{center}
\label{table:ablation_thumos_2}
\end{table}

\subsection{Comparison with state-of-the-art methods}
We compare our method with the existing fully-supervised and weakly-supervised methods under several IoU thresholds.
We separate the entries by horizontal lines regarding the levels of supervision. For readability, all results are reported on the percentage scale.

\Tref{table:quant_thumos} demonstrates the results on THUMOS'14. As shown, our method achieves a new state-of-the-art performance on weakly-supervised temporal action localization.
Notably, our method significantly outperforms the existing background modeling approaches, Liu \textit{et al.}~\cite{liu2019completeness}, Nguyen \textit{et al.}~\cite{Nguyen2019WeaklySupervisedAL} and BaS-Net~\cite{lee2020background}, by large margins of 9.5 \%, 5.6 \% and 6.6 \% in terms of average mAP, respectively.
This confirms the effectiveness of our uncertainty modeling.
Moreover, with a much lower level of supervision, our method beats several fully-supervised methods, following the latest approaches with the least gap.

The performances on ActivityNet 1.2 are demonstrated in \Tref{table:quant_anet12}. Consistent with the results on THUMOS'14, our method outperforms all weakly-supervised approaches by obvious margins, following SSN~\cite{zhao2017temporal} with a small gap less than 1 \%.
We also summarize the results on ActivityNet 1.3 in \Tref{table:quant_anet13}. We see that our method surpasses existing weakly-supervised methods including those which use external information.
Furthermore, our method performs better than the supervised method TAL-Net~\cite{chao2018rethinking} in terms of average mAP, demonstrating the potential of weakly-supervised action localization.

\begin{figure*}[t]
  \centering
  \includegraphics[clip=true, width=0.98\textwidth]{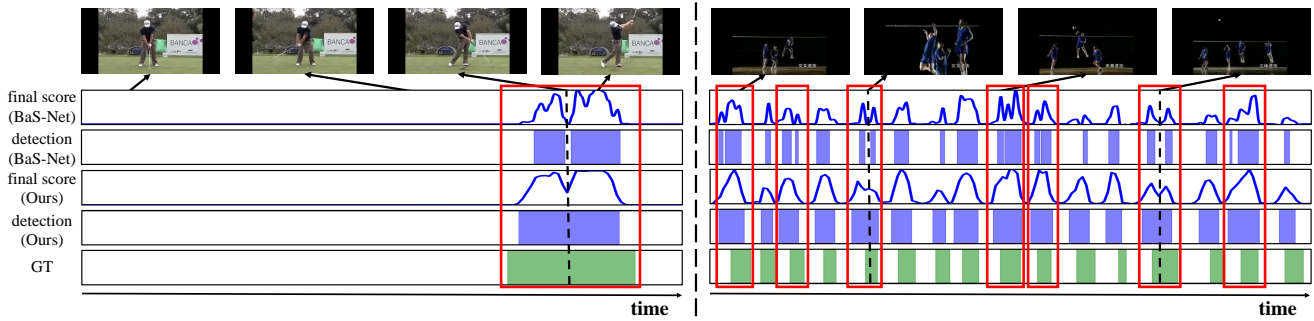}
  \caption{Qualitative comparison with BaS-Net~\cite{lee2020background} on THUMOS'14.
  We provide two different cases of examples: (1) a sparse action case with \textit{GolfSwing} and (2) a frequent action case with \textit{VolleyballSpking}.
  There are five plots with sampled frames for each example.
  The first and second plot show the final scores and the detection results of the corresponding action class from BaS-Net respectively, while the third and fourth plot represent those from our method respectively.
  The last plot is the ground truth action intervals.
  The horizontal axis in each plot means the timesteps of the video, while the vertical axes are the score values from 0 to 1.
  In the red boxes, while BaS-Net fails to cover complete action instances and splits them into multiple detection results, our method precisely localizes them.
  The black dashed lines indicate the action frames that are misclassified by BaS-Net but detected by our method.}
  \label{fig:qualitative_comparison}
\end{figure*}

\subsection{Ablation study}
\paragraph{Effects of loss components and score calculation.}
In \Tref{table:ablation_thumos_1}, we investigate the contribution of each component on THUMOS'14.
Firstly, the \textit{baseline} is set as the main pipeline only with video-level classification loss ($\mathcal{L}_{\text{cls}}$).
We try two types of score calculation methods for the baseline: (1) softmax and (2) fusion of softmax and uncertainty (as in Eq.~\ref{equ:posterior}).
For the second, as the original feature magnitudes are unconstrained, it is inappropriate to use Eq.~\ref{equ:ood_formul}.
Instead, we perform min-max normalization on feature magnitudes to estimate the probability of a segment being in-distribution.
Surprisingly, the fusion method largely improves the average mAP by 9.9 \%, as the relatively low magnitudes of background segments offset their softmax scores, which reduces false positives.
This confirms both the correlation between feature magnitudes and action/background segments and the importance of background modeling.
Thereafter, the proposed losses, \ie, uncertainty modeling loss ($\mathcal{L}_{\text{um}}$) and background entropy loss ($\mathcal{L}_{\text{be}}$), are added subsequently.
We note that background entropy loss cannot stand alone, as it is calculated with the pseudo background segments which are selected based on uncertainty.
As a result, uncertainty modeling boosts the performance to 40.2 \% in terms of the average mAP, which already achieves a new state-of-the-art.
In addition, background entropy loss further improves the performance, widening the gap with the existing methods.

\paragraph{Analysis on the maximum feature magnitude $m$.}
The maximum feature magnitude $m$ in Eq.~\ref{equ:ood_formul} determines how different the feature magnitudes from action frames and those from background frames are.
We investigate the effect of the maximum feature magnitude $m$ in \Tref{table:ablation_thumos_2}, where $m$ is altered from 10 to 250.
We notice that the performance differences are insignificant when $m$ is large enough.
Meanwhile, the performance decreases with a too small $m$, because the separation between action and background is insufficient.

\subsection{Qualitative results}
\label{sec:qulitative_comparison}
To confirm the superiority of our background modeling, we qualitatively compare our method with BaS-Net~\cite{lee2020background} which uses the background class.
In \Fref{fig:qualitative_comparison}, two examples from THUMOS'14 are illustrated: sparse and frequent action cases.
In the both cases, we see that our model detects the action instances more precisely.
More specifically, in the red boxes, it can be noticed that BaS-Net splits one action instance into multiple incomplete detection results.
We conjecture that this problem is because BaS-Net strongly forces inconsistent background frames to belong to one class, which makes the model misclassify confusing parts of action instances as the background class (See the black dashed lines).
On the contrary, our model provides better separation between action and background via uncertainty modeling instead of using a separate class.
Consequently, our model successfully localizes the complete action instances without splitting them.

\section{Conclusion}
In this work, we identified the inherent limitations of existing background modeling approaches, with the observation that background frames may be dynamic and inconsistent.
Thereafter, based on their inconsistency, we proposed to formulate background frames as out-of-distribution samples and model uncertainty with feature magnitudes.
In order to train the model to identify background frames without frame-level annotations, we designed a new architecture, where uncertainty is learned via multiple instance learning.
Furthermore, background entropy loss was introduced to prevent background segments from leaning toward any specific action class.
The ablation study verified that our uncertainty modeling and background entropy loss both are beneficial for the localization performance.
Through the experiments on two most popular benchmarks - THUMOS'14 and ActivityNet, our method achieved a new state-of-the-art with a large margin on weakly-supervised temporal action localization.
We believe it would be a promising future direction to adopt the out-of-distribution formulation of background to the fully-supervised setting or other related tasks.

\section{Acknowledgement}
% This project was partly supported by Next-Generation Information
% Computing Development Program through the National Research
% Foundation of Korea (NRF) funded by the Ministry of Science and
% ICT (NRF-2017M3C4A7069370) and the Institute for Information \& Communications Technology Planning \& Evaluation (IITP)
% grant funded by the Korea government (No. 2019-0-01558: Study
% on audio, video, 3d map and activation map generation system using deep generative model; No. 2020-0-01361: Artificial Intelligence Graduate School
% Program (YONSEI UNIVERSITY)).
This project was partly supported by the National Research Foundation
of Korea (NRF) grant funded by the Korea government
(MSIT) (No. 2019R1A2C2003760) and the Institute for Information \& Communications Technology Planning \& Evaluation (IITP) grant funded by the Korea government (No. 2019-0-01558: Study on audio, video, 3d map and activation map generation system using deep generative model; No. 2020-0-01361: Artificial Intelligence Graduate School Program (YONSEI UNIVERSITY)).

{
\bibliography{aaai21}
}

\end{document}